\documentclass[acmtog, nonacm]{acmart}

\usepackage{multirow}
\usepackage{appendix}

\begin{document}

\title{TAMF-VTON: Token-Wise Sparse Adaptation for Mask-Free Virtual Try-On}


\author{Jie Wang}
\email{wangjie@linctex.com}
\affiliation{%
  \institution{State Key Lab of CAD \& CG, Zhejiang University and Style3D Research}
  \country{China}}

\author{Qian He}
\email{heqianhailie@gmail.com}
\affiliation{%
  \institution{State Key Lab of CAD \& CG, Zhejiang University and Style3D Research}
  \country{China}}

\author{Gaofeng He}
\affiliation{%
  \institution{Style3D Research}
  \country{China}}

\author{Xiaogang Jin}
\affiliation{%
  \institution{State Key Lab of CAD \& CG, Zhejiang University}
  \country{China}}

\author{Huamin Wang}
\affiliation{%
  \institution{Style3D Research}
  \country{China}}

\begin{abstract}
Diffusion-based virtual try-on (VTON) often relies on segmentation-derived masks that become unreliable when the reference garment differs substantially from the target person's clothing. Existing methods may also lose fine details and commonly focus on single-garment transfer. We present TAMF-VTON, an inference-time mask-free framework built on Qwen-Image-Edit-2509. Given a person image, one or more garment images, and a text instruction, it performs single- or multi-garment transfer without masks, parsing maps, pose representations, or spatial controls at inference. Its central component is token-wise sparse Mixture-of-Experts adaptation, which provides input-dependent capacity without predefined expert roles. Timestep-weighted frequency supervision regularizes high-frequency reconstruction, while adaptive offline data construction supplies cross-category and cross-silhouette training tuples; neither is used at inference. TAMF-VTON achieves the best scores on most reported paired and unpaired metrics among evaluated VTON baselines on the standard single-garment VITON-HD and DressCode benchmarks. Selected qualitative results illustrate multi-garment composition, subject-to-subject transfer, diverse scenes, body-shape variation, and illumination changes. With INT4 on an RTX 4090, TAMF-VTON generates 1K-resolution images under 15 seconds and 2K-resolution images under 30 seconds. Our project can be accessed via \textit{https://www.style3d.ai/ai-photoshoot/virtual-clothing-try-on}.
\end{abstract}



\begin{teaserfigure}
    \centering
    \includegraphics[width=1.0\textwidth]{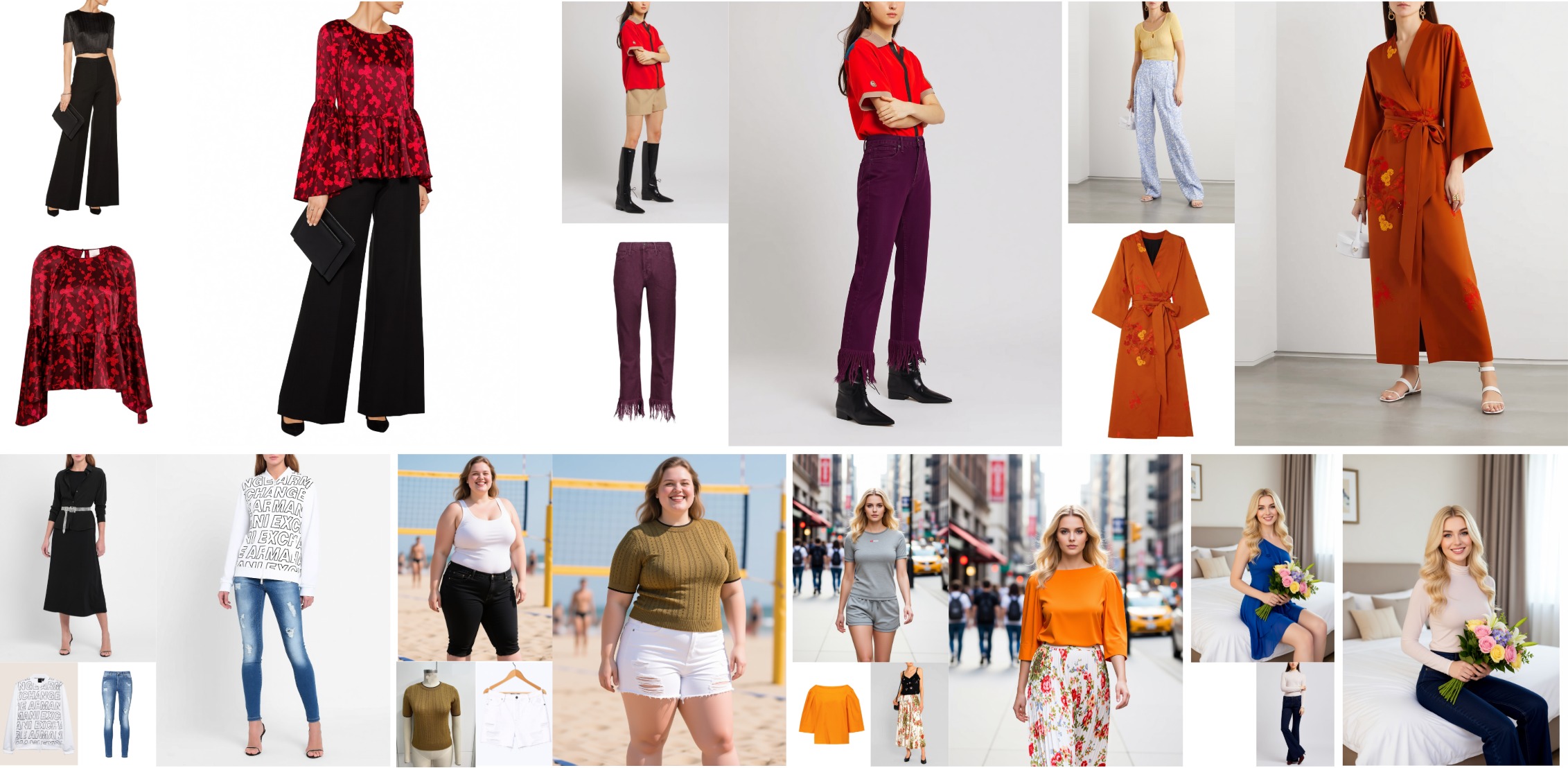}
    \caption{Qualitative results of TAMF-VTON across diverse virtual try-on scenarios. The first row shows single-garment transfer from a flat-lay product image to a studio person image. The second row demonstrates more challenging settings, including multi-garment composition, in-the-wild try-on, and subject-to-subject garment transfer. TAMF-VTON preserves non-try-on regions and exposed body parts with strong spatial consistency, maintaining the original background, person identity, and body structure while faithfully transferring fine-grained garment details.}
    \label{fig1}
\end{teaserfigure}

\maketitle

\section{Introduction}

Recent diffusion-based generative models have advanced text-to-image generation and image-to-image translation~\cite{rombach2022, sd3, podell2024sdxl, labs2025}. Among their applications, virtual try-on (VTON) aims to synthesize a realistic image of a target person wearing one or more reference garments while preserving identity, pose, body structure, and scene context. This task is increasingly important for e-commerce and digital fashion.

Despite rapid progress, existing diffusion-based VTON methods remain limited in less constrained settings. Mask-based approaches~\cite{Zhu_2023_CVPR_tryondiffusion, Kim2024_CVPR, Kim2025_ICCV, li2024, jiang2024, DBLP:journals/corr/abs-2406-04542, zhou_2025_CVPR} formulate VTON as inpainting and use segmentation masks, human parsing maps, or related spatial priors to define the editable region. Errors in these inputs can propagate to the output under complex poses, occlusion, loose clothing, or cross-category transfer. For example, a region derived from a short T-shirt may not cover the extent required by a long coat. Recent mask-free methods~\cite{zhang_boow_vton, wan_11375578, Du_2025_ICCV} instead learn garment--person alignment from image pairs. However, many of these methods require garment-specific branches or extensive backbone adaptation, and most are designed primarily for single-garment transfer.

We formulate VTON as task-specialized image editing and ask whether input-dependent sparse updates can adapt a frozen multimodal editor more effectively than one shared low-rank update. Here, \emph{mask-free} refers specifically to inference: the model receives only person and garment images plus text. Segmentation, pose, and masks are restricted to offline data construction. TAMF-VTON combines token-wise sparse MoE adaptation of Qwen-Image-Edit-2509 with frequency supervision and adaptive offline data construction.

Our contributions are threefold:
\begin{itemize}
    \item We introduce token-wise sparse adaptation of a frozen Qwen-Image-Edit-2509 backbone, routing each multimodal token to a subset of low-rank experts without predefined expert roles.

    \item We combine timestep-weighted frequency supervision with adaptive offline construction of cross-category and cross-silhouette training tuples, without adding inference-time controls.

    \item We quantitatively evaluate paired and unpaired single-garment transfer on VITON-HD and DressCode, and qualitatively assess multi-garment and challenging-scene capabilities against accessible specialized and general-purpose baselines.
\end{itemize}

\section{Related Work}
\subsection{GAN-based Virtual Try-On}

Early image-based virtual try-on methods were mainly built upon generative adversarial networks (GANs) and commonly followed a two-stage pipeline: first deforming the reference garment to match the target pose, and then blending it with the person image~\cite{issenhuth2020, men2020, bai2022_eccv}. VITON~\cite{han2018} introduced a coarse-to-fine framework with thin-plate spline (TPS) transformation for garment warping, while PF-AFN~\cite{ge2021} improved alignment through appearance flow and knowledge distillation. Subsequent works further incorporated attention mechanisms or refined flow estimation to enhance geometric fidelity~\cite{lee2022}. Despite these improvements, GAN-based methods remain sensitive to inaccurate warping and often struggle to synthesize photorealistic results with faithful preservation of fine-grained textures, such as patterns, fabric weaves, and material details.

\subsection{Mask-based Virtual Try-On}
Recent advances in diffusion models have catalyzed a new generation of high-fidelity virtual try-on (VTON) methods, most of which cast the task as reference-guided image inpainting. TryonDiffusion \cite{Zhu_2023_CVPR_tryondiffusion}, StableVITON~\cite{Kim2024_CVPR}, LaDI-VTON~\cite{morelli2023ladi}, DCI-VTON~\cite{gou2023taming}, OOTDiffusion \cite{xu2025}, IDM-VTON \cite{choi2024}, FitDiT \cite{jiang2024}, and DreamFit \cite{lin2025} adopt a dual-branch architecture, where a ReferenceNet encodes the appearance and structure of the reference garment into features that are injected into a DenoisingNet via cross-attention. In contrast, concatenation-based approaches such as CatVTON \cite{chong2025, chong2025_v2} and Voost \cite{lee2025} avoid duplicating the diffusion backbone by directly concatenating the target person and garment images as input to a single denoising network. M\&M VTON \cite{DBLP:journals/corr/abs-2406-04542} leverages a DiT backbone to enable multi-garment VTON by concatenating the tokens of the reference garments and target person. 

Despite their strong visual quality, masked diffusion-based methods inherit several limitations from the inpainting formulation. First, the inpainting mask is typically derived from the target person's original clothing, which may not match the category, shape, or spatial extent of the reference garment. This mismatch becomes particularly problematic in cross-category try-on and multi-garment composition. Second, the inpainting mask indiscriminately erases not only the original garment but also underlying body structure and background context, compromising body shape and scene consistency. Third, these methods fine-tune off-the-shelf text-to-image diffusion models. This scheme prioritizes global coherence over local texture-specific fidelity. Furthermore, fine-tuning the diffusion backbone incurs substantial computational overhead, consumes extensive training pairs, and compromises the generality of the base model.

\subsection{Mask-Free Virtual Try-On}

To overcome the dependency on explicit masks, recent works have explored mask-free VTON settings. BOOW-VTON~\cite{zhang_boow_vton}, MFT-VITON~\cite{wan_11375578}, and All Parts Matter~\cite{Du_2025_ICCV} learn garment-person alignment directly from image pairs, reducing the need for manually specified inpainting regions. These methods improve input flexibility and alleviate artifacts caused by inaccurate masks. However, existing mask-free methods still face important constraints. Many of them rely on additional garment-specific branches or large-scale backbone fine-tuning, which increases training and inference cost. Moreover, most are designed primarily for single-garment transfer and provide limited support for flexible multi-garment composition. Their training data construction also remains challenging, especially when the synthesized region must be semantically consistent with the target garment category.

In contrast, TAMF-VTON formulates virtual try-on as a task-specialized image editing problem rather than a mask-guided inpainting task. Our framework requires no segmentation masks, human parsing, or auxiliary control signals during inference. Instead of introducing a separate garment encoder or fine-tuning the full diffusion backbone, we propose a lightweight MoE adaptation scheme with dynamic expert routing to inject try-on-specific expertise while preserving the general editing capability of the pretrained model. By using token-wise concatenation, TAMF-VTON naturally supports both single-garment transfer and flexible multi-garment composition. We further introduce an adaptive inpainting-based data curation strategy for semantically aligned pseudo-pair construction, together with frequency-domain supervision to improve high-frequency texture consistency in the synthesized results.

\section{Preliminaries}
Flow matching \cite{lipman2023} formulates the generative process as learning a continuous velocity field that transports samples from a noise distribution to the data manifold along straight-line trajectories. Given an image \(x_0\), its latent representation \(z_0\) and noisy latent \(z_t\) at time \(t \in [0,1]\) are defined as:
\begin{equation}
z_0 = \mathcal{E}(x_0),
\end{equation}
\begin{equation}
    z_t = (1 - t) z_0 + t \epsilon,
\end{equation}
where \(\mathcal{E}\) is a pre-trained VAE encoder and \(\epsilon \sim \mathcal{N}(0,I)\) denotes a random noise sample. The velocity \(v_t\) at time \(t\) is given by:
\begin{equation}
    v_t = \frac{dz_t}{dt} = \epsilon - z_0.
\end{equation}

The flow model \(v_\theta\) (parameterized by \(\theta\)) takes as input the noisy latent \(z_t\), time \(t\) and conditioning signal \(c\) to predict the velocity \(v_t\). The model is optimized using a weighted mean squared error loss:
\begin{equation}
    \mathcal L_{\mathrm{FM}} = \mathbb{E}_{\epsilon \sim \mathcal{N}(0,I), t \sim \mathcal{U}(0, 1)}[\omega(t)\|v_{\theta}(z_t,t,c) - v_t\|_2^2],
\end{equation}
where \(\omega(t)\) is a positive weighting function, \(\mathcal{N}(0,I)\) denotes the standard isotropic Gaussian distribution, and \(\mathcal{U}(0, 1)\) is the uniform distribution over \([0, 1]\).

\section{Method}

\begin{figure*}
    \centering
    \includegraphics[width=1.0\textwidth]{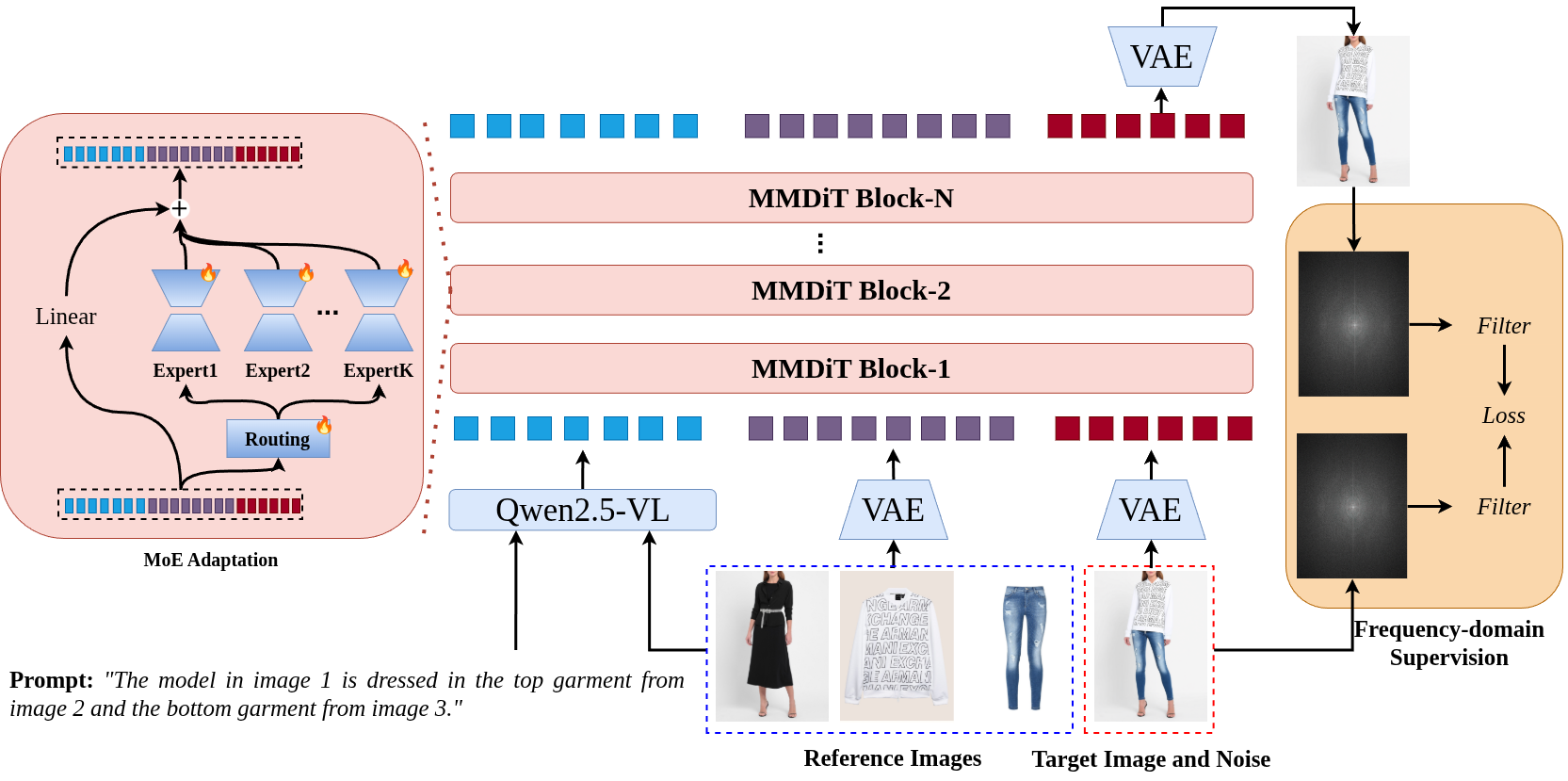}
    \caption{Overview of the TAMF-VTON architecture. Multimodal inputs are encoded as visual and textual tokens, concatenated, and processed by the MoE-augmented Qwen-Edit backbone (left). Frequency-domain supervision minimizes high-frequency spectral discrepancies during training (right). Only the MoE routing and expert parameters are updated, while the original backbone remains frozen.}
    \label{fig2}
\end{figure*}

As illustrated in Figure~\ref{fig2}, TAMF-VTON adopts Qwen-Edit~\cite{wu2025} for mask-free virtual try-on. Given a target person image, one or more reference garment images, and a text prompt, Qwen-VL extracts textual tokens, while the VAE encoder converts the target and reference images into latent visual tokens. These tokens are concatenated into a unified multimodal sequence and fed into the MMDiT backbone to predict the flow-matching velocity field.

TAMF-VTON introduces three key designs: a lightweight MoE adaptation module inserted into each linear layer for efficient task-specific specialization, a frequency-domain supervision objective for high-frequency texture preservation, and an adaptive inpainting strategy for scalable pseudo-pair construction. We describe these designs in Sections~\ref{moe}, \ref{frequency}, and~\ref{data}, respectively.

\subsection{Token-Wise Sparse MoE Adaptation}\label{moe}
Virtual try-on forms a heterogeneous multimodal sequence: visual tokens come from the target person and separate garment references with different spatial content, while text tokens specify the requested edit. A conventional LoRA module~\cite{hu2022} applies a single shared low-rank update to all tokens, which may provide insufficient conditional capacity for the diverse appearance and geometric variations encountered in VTON. We therefore introduce token-wise sparse MoE adaptation. Each token dynamically activates a subset of LoRA experts, providing input-dependent adaptation capacity while retaining sparse computation. The experts are not assigned predefined semantic or task-specific roles.

Unlike ICEdit~\cite{zhang2025icedit}, which targets general instruction-driven editing, and TT-LoRA~\cite{kunwar_ttlora}, which routes at the task level, our router operates on individual multimodal tokens. It does not use predefined task labels, parsing priors, or handcrafted expert assignments. We interpret this mechanism as input-dependent sparse capacity rather than assigning semantic functions, such as geometry or texture processing, to individual experts.

Given a token feature \(x \in \mathbb{R}^{d_{\mathrm{in}}}\), a linear router produces a distribution over \(K\) experts:
\begin{equation}
    g(x) = \operatorname{softmax}(W_r x + b_r),
\end{equation}
where \(W_r \in \mathbb{R}^{K \times d_{\mathrm{in}}}\) and \(b_r \in \mathbb{R}^{K}\) are learnable router parameters. Let \(\mathcal{T}_k(x)\) contain the indices of the \(k\) largest entries of \(g(x)\). After top-\(k\) selection, we renormalize the retained weights:
\begin{equation}
    \widetilde{g}_i(x) =
    \frac{g_i(x)}{\sum_{j \in \mathcal{T}_k(x)} g_j(x)},
    \qquad i \in \mathcal{T}_k(x).
\end{equation}
The corresponding MoE-augmented linear projection is
\begin{equation}
    y = Wx + \frac{\alpha}{r}
    \sum_{i \in \mathcal{T}_k(x)}
    \widetilde{g}_i(x) B_i A_i x,
\end{equation}
where \(W \in \mathbb{R}^{d_{\mathrm{out}} \times d_{\mathrm{in}}}\) is the frozen backbone projection, \(A_i \in \mathbb{R}^{r \times d_{\mathrm{in}}}\) and \(B_i \in \mathbb{R}^{d_{\mathrm{out}} \times r}\) are the parameters of the \(i\)-th LoRA expert, \(r\) is the LoRA rank, and \(\alpha\) is the LoRA scaling factor. We use \(K=4\) total experts, activate \(k=2\) experts per token, and set \(r=64\) and \(\alpha=32\), giving \(\alpha/r=1/2\).

During training, we freeze the pretrained diffusion backbone and optimize only the routers and LoRA experts. Sparsity is imposed solely by top-\(k\) selection. The resulting module increases conditional adaptation capacity while limiting the number of experts activated for each token.

\subsection{Frequency-Domain Supervision for Textural Detail Consistency}\label{frequency}

\begin{figure}[t]
    \centering
    \includegraphics[width=0.5\textwidth]{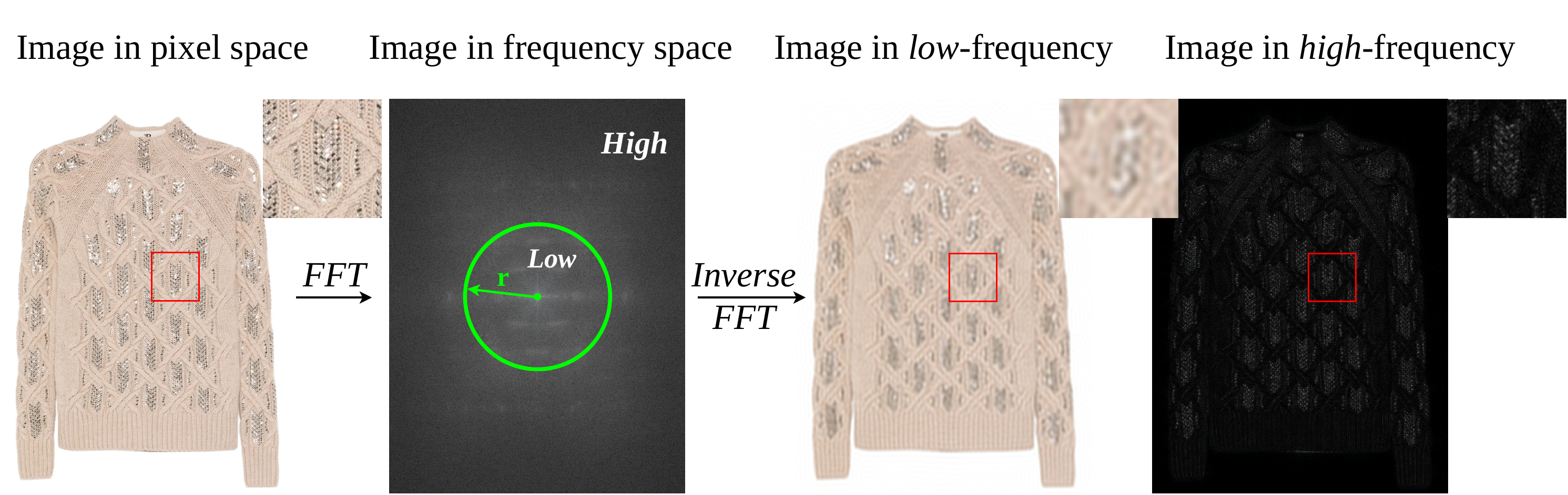}
    \caption{Visualization of garment representations in the pixel and frequency domains. Inverse FFT reconstruction from low- and high-frequency bands shows that fine-grained texture details are predominantly encoded in high-frequency components.}
    \label{fig3}
\end{figure}

Fine-grained garment textural details, such as fabric grain, weave structures, logos, and local patterns, are often attenuated during diffusion-based synthesis. Since these details are primarily reflected in the high-frequency components of the Fourier spectrum, we introduce a frequency-domain supervision mechanism to explicitly encourage spectral consistency between the synthesized result and the ground-truth image. Unlike prior frequency-based losses FFL~\cite{jiang2021focal} for image reconstruction and synthesis, our method is designed specifically for diffusion-based mask-free VTON, where preserving globally consistent garment textures is critical under large geometric deformation and cross-category transfer. As illustrated in Figure~\ref{fig3}, in the FFT-shifted spectrum, low-frequency components near the center mainly capture coarse structure and color distribution, while peripheral high-frequency regions encode fine textural details.

\subsubsection{Spectral Decomposition and High-Frequency Loss}
Given the prediction in latent space, we first reconstruct an estimate of the clean image \(x_0\) to enable spectral analysis. Under the flow matching formulation, the model predicts the velocity field \(\hat{v}_t = v_\theta(z_t,t,c)\). Since the ground-truth velocity is \(v_t = \epsilon - z_0\). Inverting this relation yields an estimate of the data latent:
\begin{equation}
    \hat{z}_0 = z_t - t\hat{v}_t.
\end{equation}

The estimated latent is then decoded into pixel space using the frozen VAE decoder: \(\hat{x}_0 = \mathcal{D}(\hat{z}_0)\). Next, we apply the FFT to both \(\hat{x}_0\) and the ground-truth image \(x_0\), obtaining their spectral representations \(\mathcal{F}(\hat{x}_0)\) and  \(\mathcal{F}(x_0)\). To focus on texture-related frequency components, we construct a radial high-frequency mask \(M_f\) on the FFT-shifted spectrum:
\begin{equation}
    M_f(u, v) = 
    \begin{cases}
    1, & \text{if } \sqrt{u^2 + v^2} > r, \\
    0, & \text{otherwise},
    \end{cases}
\end{equation}
where \(r=(h+w)/32\) denotes a resolution-related cutoff radius and \((u,v)\) denotes the centered frequency coordinate.

Different from masked frequency supervision used in prior VTON method FitDiT ~\cite{jiang2024}, we apply spectral constraints on the reconstructed full image rather than local garment regions. This is motivated by the fact that Fourier coefficients are globally coupled, such that spatial masking in the pixel domain does not correspond to localized frequency decomposition. Consequently, enforcing spectral consistency on masked regions may introduce inconsistent frequency constraints under large garment deformation or cross-category transfer.

Moreover, the reconstructed image \(\hat{x}_0\) becomes less reliable at high-noise timesteps. Directly imposing strong spectral supervision in such stages may therefore destabilize optimization. To address this issue, we introduce a timestep-dependent weighting term:
\begin{equation}
\gamma(t) = (1-t)^2,
\end{equation}
which emphasizes frequency supervision at low-noise stage and suppresses it when \(t\) is close to the noise endpoint.

The final high-frequency spectral loss is formulated as:
\begin{equation}
    \mathcal{L}_{f}
    =
    \gamma(t)
    \cdot
    \frac{1}{\|M_f\|_1}
    \left\|
    M_f \odot
    \left(
    \mathcal{F}(\hat{x}_0) - \mathcal{F}(x_0)
    \right)
    \right\|_2^2,
\end{equation}
where \(\odot\) denotes element-wise multiplication.

\subsubsection{Unified Training Objective}
Our overall objective combines the standard flow matching loss \(\mathcal L_{\mathrm{FM}}\) with the proposed frequency-aware term \(\mathcal L_f\):
\begin{equation}
    \mathcal{L} = \mathcal L_{\mathrm{FM}} + \lambda\mathcal L_f,
\end{equation}
where \(\lambda\) controls the contribution of the spectral term. This diffusion-aware frequency supervision provides an effective inductive bias for preserving fine-grained garment textures while remaining fully compatible with mask-free VTON.

\subsection{Adaptive Inpainting Strategy for Mask-Free Data Curation}\label{data}

\begin{figure}[t]
    \centering
    \includegraphics[width=0.5\textwidth]{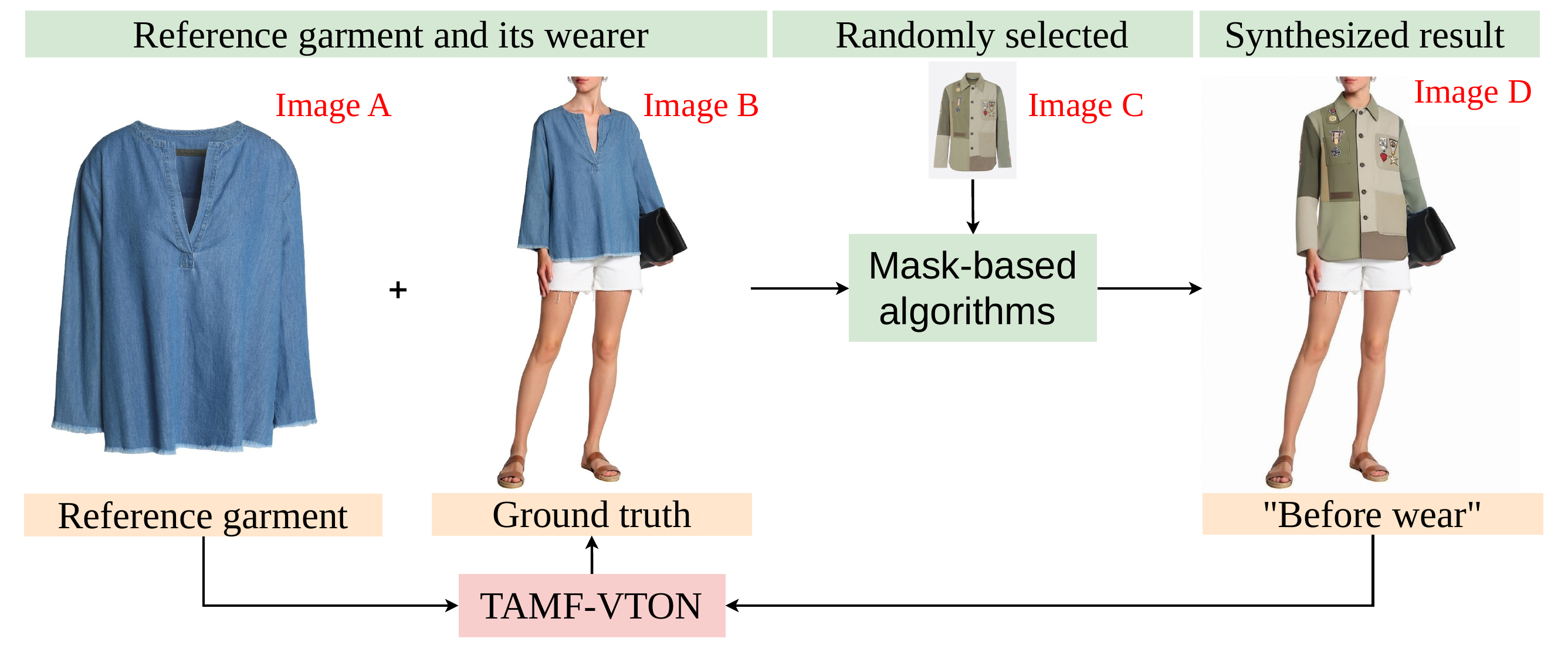}
    \caption{Pipeline for synthesizing mask-free VTON training samples. We first generates a reference-agnostic base image with an adaptive inpainting prior, and then construct aligned training tuples for mask-free try-on training.}
    \label{fig4}
\end{figure}

Training a mask-free VTON model requires aligned training tuples consisting of a reference-agnostic person image, one or more reference garment images, and the corresponding ground-truth try-on image. However, existing datasets such as VITON-HD~\cite{choi2021} and DressCode~\cite{morelli2022} usually provide only paired person-garment images, without pixel-aligned "before-wear" images. We therefore synthesize mask-free training tuples through an inverse VTON process, as illustrated in Figure~\ref{fig4}.

Given a target person image \(I_{\mathrm{target}}\) (Image B) wearing garment \(G_{\mathrm{ref}}\) (Image A), we first construct a reference-agnostic base image \(I_{\mathrm{base}}\) (Image D) by replacing the original garment with a randomly sampled garment (Image C) using a mask-based VTON model. The generated \(I_{\mathrm{base}}\) preserves the target identity, pose, and background while removing garment-specific appearance from \(G_{\mathrm{ref}}\). We then rearrange the samples into a mask-free training triplet \((I_{\mathrm{base}}, G_{\mathrm{ref}}, I_{\mathrm{target}})\), where the model learns to recover the original target image from the agnostic base and the reference garment. The editing instruction for single-garment is: \textit{"The model in image 1 is dressed in the top/bottom/dress garment from image 2"}. This construction naturally extends to multi-garment and subject-to-subject settings by using multiple reference garments or donor-model garments.

\subsubsection{Adaptive Inpainting Strategy}

A key limitation of prior mask-free data pipelines~\cite{zhang_boow_vton, wan_11375578} is that their inpainting masks are usually built from the union of human parsing masks and the target person's original garment mask. Such masks are tied to the originally worn clothing and often fail when the reference garment has a different category, length, or silhouette, such as replacing a T-shirt with a long coat. To generate more suitable agnostic bases for cross-category and multi-garment try-on, we propose an adaptive inpainting strategy that estimates the inpainting region according to both reference-garment semantics and target-person pose.

\begin{figure}[t]
    \centering
    \includegraphics[width=0.5\textwidth]{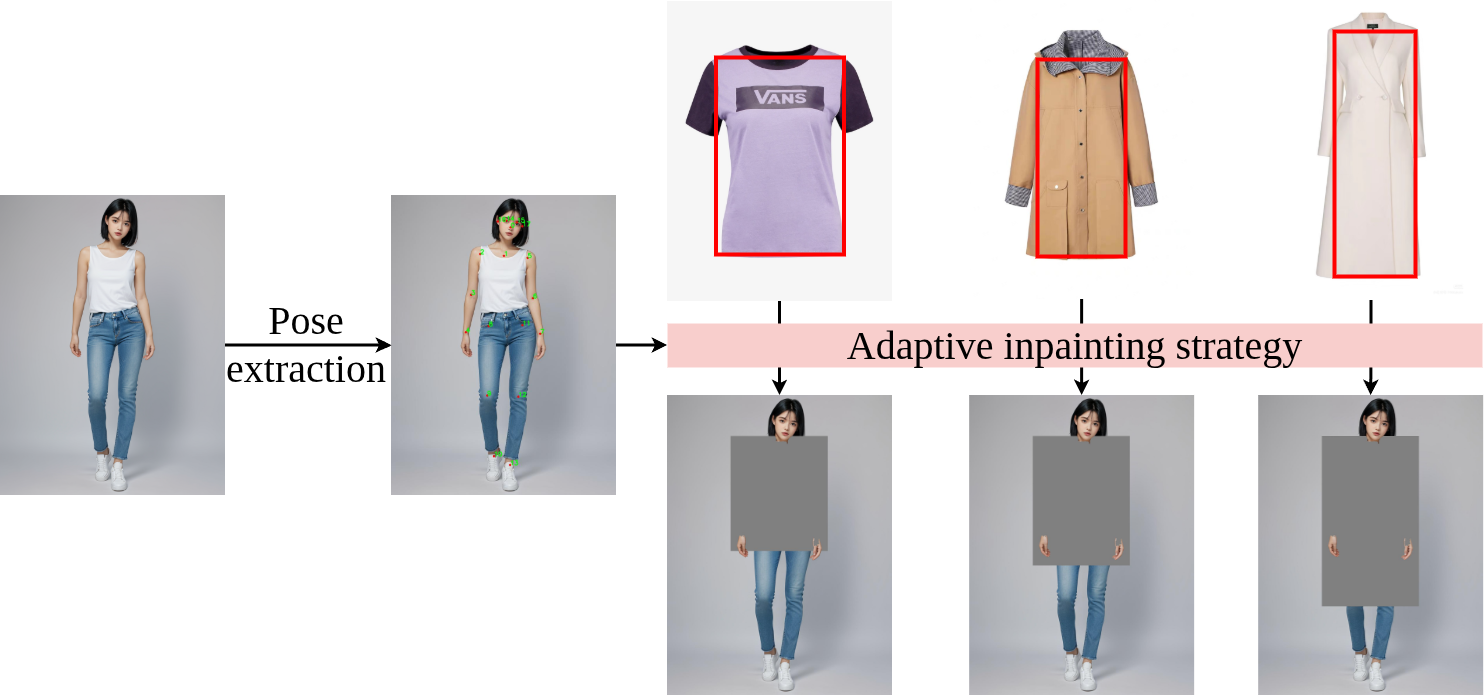}
    \caption{Adaptive inpainting mask construction. The red rectangle indicates the detected region used for semantic-aware mask prediction. The adaptive inpainting mask is shown as the gray rectangle.}
    \label{fig5}
\end{figure}

Specifically, we detect the main garment region in \(G_{\mathrm{ref}}\) (red rectangle in Figure~\ref{fig5}) using a YOLOv10-based detector~\cite{wang2024}, obtaining its category and height-to-width ratio as coarse cues for garment type and length. We also extract normalized 2D body keypoints using OpenPose~\cite{cao2019} to capture pose-dependent body layout. These cues are fed into a lightweight linear regressor to predict the garment hemline position in the target image, from which we construct an adaptive rectangular inpainting region. The final agnostic mask is the union of this predicted region and the original garment mask, ensuring coverage of both the expected reference-garment footprint and residual textures from the original clothing.

We further protect hands and feet using SAM2 segmentations~\cite{ravi2024} during inpainting, reducing limb distortion caused by overly large masks. We recompute agnostic masks for VITON-HD~\cite{choi2021} and DressCode~\cite{morelli2022}, and retrain FitDiT~\cite{jiang2024} with the updated masks to obtain a VTON-oriented inpainting prior. Compared with pipelines that rely on fixed parsing-based masks or chained multimodal models, our strategy provides a simple and scalable way to synthesize high-quality aligned training tuples for mask-free single- and multi-garment VTON. Importantly, the synthesized samples are not constrained by style, category, or length similarity between the reference garments and the garments originally worn by the target person, enabling diverse cross-category and cross-silhouette training examples.

\section{Experiments}

\subsection{Experimental Setup}
We train TAMF-VTON on two complementary datasets. The first is a synthetic benchmark dataset constructed from VITON-HD~\cite{choi2021} and DressCode~\cite{morelli2022} using our data curation pipeline. It contains 20K balanced samples at \(768 \times 1024\) resolution across three categories: upper-body, lower-body, and dresses. The second is a 2K high-resolution dataset at \(1536 \times 2048\), synthesized from in-the-wild commercial fashion images to improve real-world generalization and texture fidelity.

Training follows a two-stage schedule: 10K steps on the 20K set and 5K fine-tuning steps on the 2K high-resolution set. We use a total batch size of 4 on NVIDIA A100 GPUs, FP8 quantization, AdamW8bit with learning rate \(5 \times 10^{-5}\), and a MoE adapter with rank 64, 4 experts, and top-2 routing. At inference, we use the Nunchaku acceleration engine~\cite{li2025svd}; TAMF-VTON produces one result under 15 seconds on an NVIDIA RTX 4090.

\subsection{Qualitative Results}
\subsubsection{Standard Benchmarks Evaluation}

We first evaluate TAMF-VTON on VITON-HD~\cite{choi2021} and DressCode~\cite{morelli2022}, which mainly involve garment transfer from flat-lay product images to studio-captured person images. We presents both single-garment and multi-garment try-on comparisons against three recent SOTA open-source baselines ~\cite{wan_11375578, zhang_boow_vton, Du_2025_ICCV}. For multi-garment transfer, the compared methods are applied sequentially to the upper- and lower-body garments.

As shown in the Figure~\ref{fig6}, TAMF-VTON generates more realistic results with accurate garment geometry, natural draping, and superior preservation of identity and body structure. Our method also better retains fine-grained textures and material appearance, while existing approaches often produce blurred or hallucinated patterns, modify non-garment regions, or distort body contours near the transferred garments. Notably, mask-based pipelines are highly sensitive to segmentation errors, whose accumulated artifacts often prevent them from achieving product-level visual quality in practical scenarios. The advantage of TAMF-VTON becomes more pronounced in multi-garment settings, where competing methods frequently exhibit inconsistent garment layering, structural artifacts, or degraded texture fidelity after sequential editing.

\subsubsection{Challenging Real-World Scenarios}

We further evaluate TAMF-VTON on scenarios beyond standard benchmarks, including multi-garment composition, complex backgrounds, and subject-to-subject transfer. As shown in the second row of Figure~\ref{fig1}, our method handles these challenging cases robustly, producing coherent garment layering, realistic textural appearance, and strong background preservation. Additional results are shown in Figures~\ref{fig9} and \ref{fig10}. These results demonstrate that TAMF-VTON generalizes well beyond controlled benchmark settings to more practical real-world scenarios.

\subsection{Quantitative Results}

\begin{table*}
    \caption{Quantitative results on VITON-HD \cite{choi2021} and DressCode \cite{morelli2022} datasets. We present paired and unpaired evaluation results. Our proposed method consistently outperforms most of the baselines. Bold indicates the best metrics in this table.}
    \label{tab1}
    \centering
    \resizebox{1.0\textwidth}{!}{
    \begin{tabular}{ccccccccccccccccc}
        \toprule
         \multirow{3}{*}{Methods} &  &  \multicolumn{7}{c}{VITON-HD \cite{choi2021}} &  & \multicolumn{7}{c}{DressCode \cite{morelli2022}}\\
         \cline{3-9}
         \cline{11-17}
         &  &  \multicolumn{4}{c}{Paired}&  &  \multicolumn{2}{c}{Unpaired}&  & \multicolumn{4}{c}{Paired}& & \multicolumn{2}{c}{Unpaired}\\
         \cline{3-6}
         \cline{8-9}
         \cline{11-14}
         \cline{16-17}
 & & SSIM $\uparrow$ & LPIPS $\downarrow$ & FID $\downarrow$ & KID $\downarrow$ & & FID $\downarrow$ & KID $\downarrow$ & & SSIM $\uparrow$ & LPIPS $\downarrow$ & FID $\downarrow$ & KID $\downarrow$ & & FID $\downarrow$ &KID $\downarrow$\\
    \hline
         IDM-VTON \cite{choi2024} &  &  0.881&  0.079&  6.338&  1.322&  &  9.611&  1.639&  & 0.923& 0.048& 3.800& 1.201& & 5.616&1.554\\
         FitDiT \cite{jiang2024} &  &  0.898&  0.066&  4.731&  \textbf{0.189} &  &  8.204&  0.342&  & 0.926& 0.043& 2.638& 0.499& & 4.732&0.901\\
         OOTDiffusion \cite{xu2025} &  &  0.851&  0.096&  6.519&  0.896&  &  9.673&  1.206&  & 0.897& 0.072& 3.950& 0.720& & 6.702&1.863\\
         CatVTON \cite{chong2025} &  &  0.869&  0.097&  6.139&  0.964&  &  9.143&  1.267&  & 0.901& 0.070& 3.275& 0.670& & 5.422&1.550\\
         Voost \cite{lee2025} & & 0.898 & 0.056 & 5.269 & 0.404 & & 8.982 &0.899 & & 0.933 & 0.044 & 2.787 &0.377 & & 5.081 & 0.787\\
         \midrule
         MFT-VITON \cite{wan_11375578} & & 0.886 & 0.088 & - & - & & 8.441 & 0.560 & & \textbf{0.939} & 0.041 & - & - & & 11.184 & 1.107 \\
         Boow-VTON \cite{zhang_boow_vton} & & 0.862 & 0.108 & 6.885 & 1.366 & & 8.809 & 0.818 & & 0.896 & 0.083 & 9.348 & 1.624 & & 11.667 & 1.292 \\
         All Parts Matter \cite{Du_2025_ICCV} & & 0.901 & 0.079 & - & - & & 9.384 & 1.120 & & 0.939 & 0.048 & - &- & & 10.950 & 1.78 \\
         \midrule
         TAMF-VTON (single-LoRA) & & 0.893 & 0.108 & 5.234 & 0.675 & & 8.341 & 0.460 & & 0.899 & 0.072 & 3.659 & 0.438 & & 4.901 & 0.942 \\
         TAMF-VTON (w/o \(\mathcal{L}_f\))& & 0.902 & 0.056 & 4.513 & 0.354 & & 7.395 & 0.289 & & 0.929 & 0.047 &  2.830 & 0.375 & & 3.633 & 0.643 \\
         TAMF-VTON &  &  \textbf{0.913} &  \textbf{0.052} &  \textbf{4.322} &  0.277&  &  \textbf{6.268} &  \textbf{0.265} &  & 0.933 & \textbf{0.040} & \textbf{2.363}& \textbf{0.360}& & \textbf{3.321}& \textbf{0.597}\\ 
         \bottomrule
    \end{tabular}
    }
    
\end{table*}

We quantitatively evaluate TAMF-VTON on the test sets of VITON-HD~\cite{choi2021} and DressCode~\cite{morelli2022}. As shown in Table~\ref{tab1}, our method achieves the best or competitive performance across most metrics, outperforming both mask-based and mask-free state-of-the-art methods. The improved SSIM and LPIPS indicate better structural preservation and perceptual similarity, while lower FID and KID demonstrate stronger photorealism and distributional consistency.

In the unpaired setting, where ground-truth outputs are unavailable and evaluation relies on distributional metrics, TAMF-VTON achieves substantially lower FID and KID than prior methods. This improvement is particularly important for cross-category and cross-style try-on, where the reference garment may differ significantly from the originally worn clothing in length, silhouette, or category. These results suggest that TAMF-VTON better decouples person structure from garment appearance, enabling more realistic synthesis under diverse try-on conditions.

Additional comparisons with recent strong baselines (e.g., Qwen-Edit-2509 ~\cite{wu2025}, Flux2-klein-9b ~\cite{flux-2-2025}, and Nano Banana Pro ~\cite{geminiteam2025geminifamilyhighlycapable}) are provided in the supplementary material.

\section{Ablation Study}
\subsection{Effect of MoE Adaptation}

\begin{figure}
    \centering
    \includegraphics[width=0.5\textwidth]{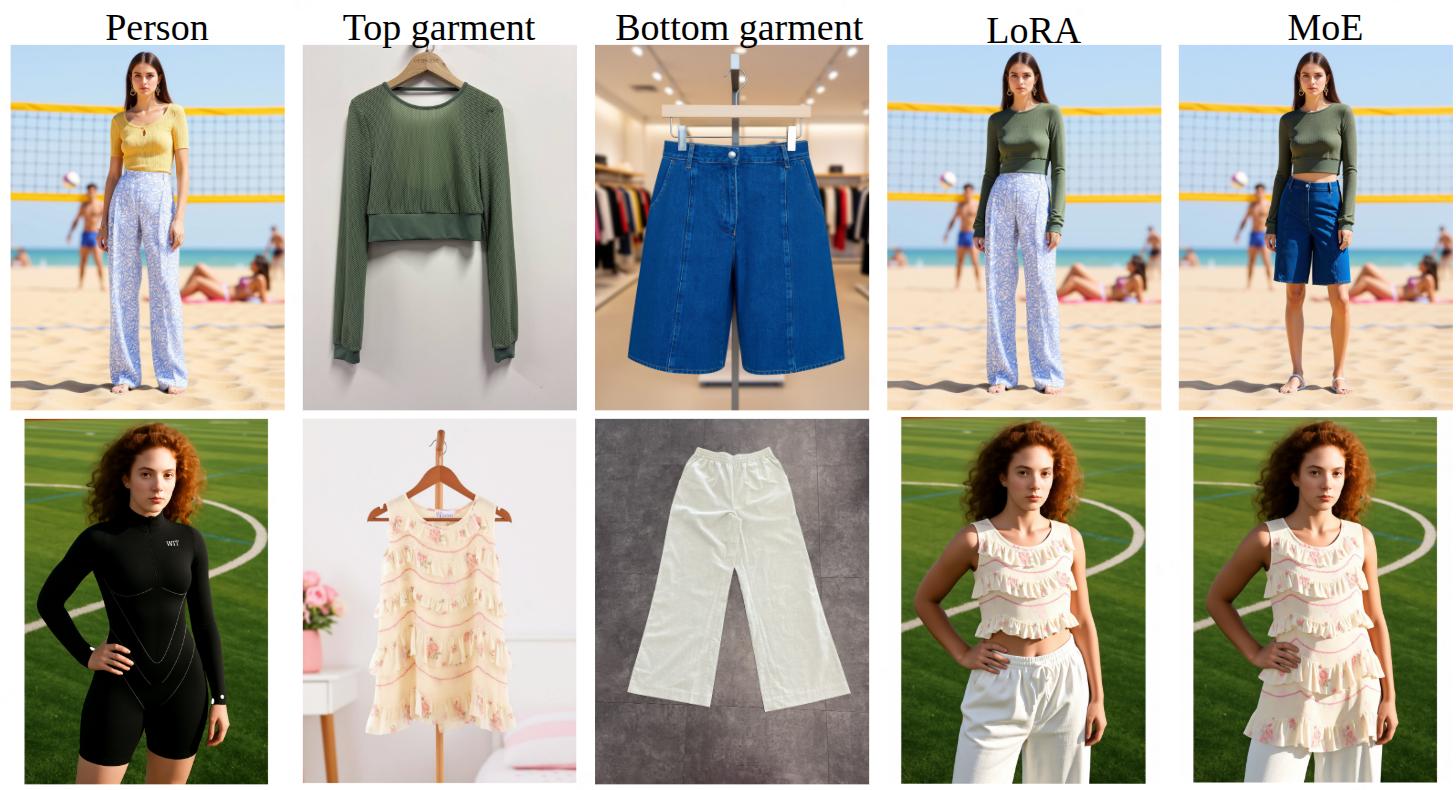}
    \caption{Ablation study on MoE adaptation. Compared with the LoRA baseline, the proposed MoE design better preserves garment structure, style cues. Best viewed zoomed in.}
    \label{fig7}
\end{figure}

To evaluate the effectiveness of the proposed Mixture-of-Experts (MoE) adaptation, we compare it with the LoRA baseline, where the MoE module is replaced by a standard LoRA adapter while keeping the training data, optimizer, training steps, and all other components unchanged. As shown in Figure~\ref{fig7}, the MoE variant produces more stable and realistic results across challenging cases. In contrast, the single-LoRA baseline shows limited generalization ability: in the first row, it fails to transfer the bottom garment; in the second row, it distorts the upper-garment silhouette and loses important style cues. Quantitative results in Table~\ref{tab1} further show that MoE adaptation consistently improves performance over the single-LoRA baseline across evaluation metrics. These results indicate that token-wise expert routing provides more effective conditional adaptation than a single shared adapter, which is particularly important for complex cases involving cross-category or multi-garment transfer.

\subsection{Effect of Frequency-Domain Supervision}
\begin{figure}
    \centering
    \includegraphics[width=0.5\textwidth]{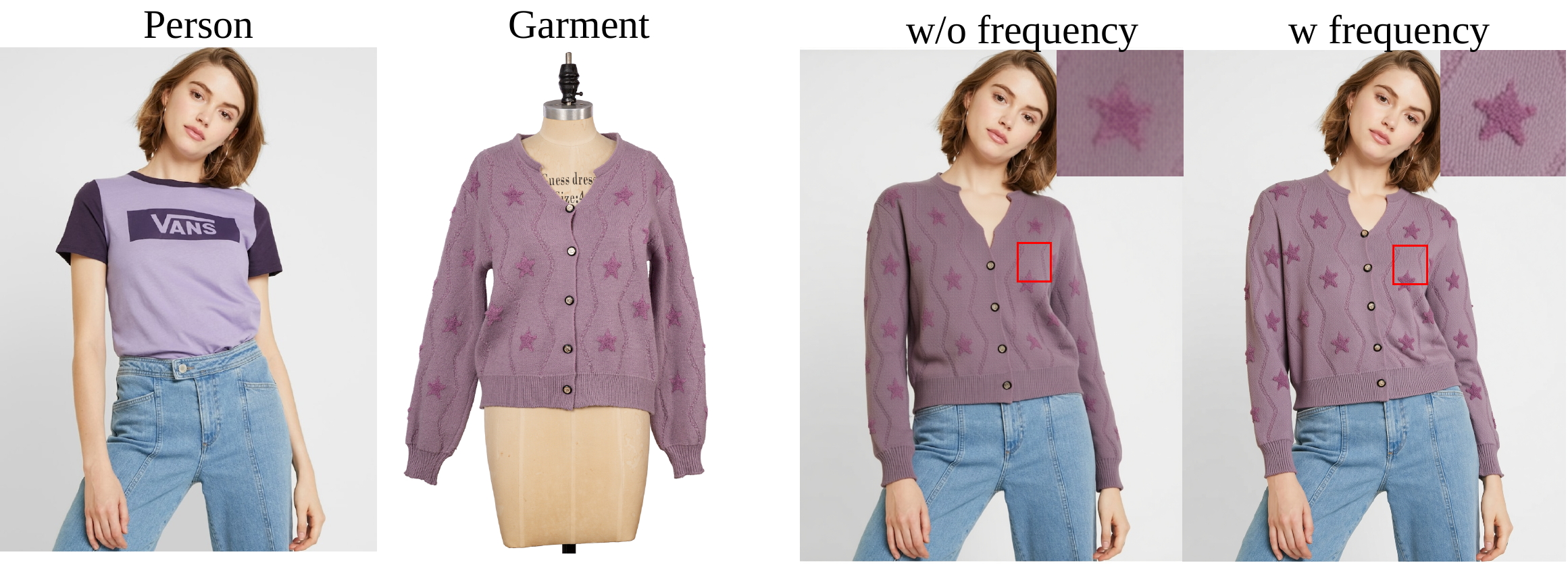}
    \caption{Ablation study on frequency-domain supervision. Removing the proposed spectral loss leads to oversmoothed or distorted local textures, while the full model better preserves fine-grained garment patterns and fabric details. Best viewed zoomed in.}
    \label{fig8}
\end{figure}

\begin{figure*}
    \centering
    \includegraphics[width=1.0\textwidth]{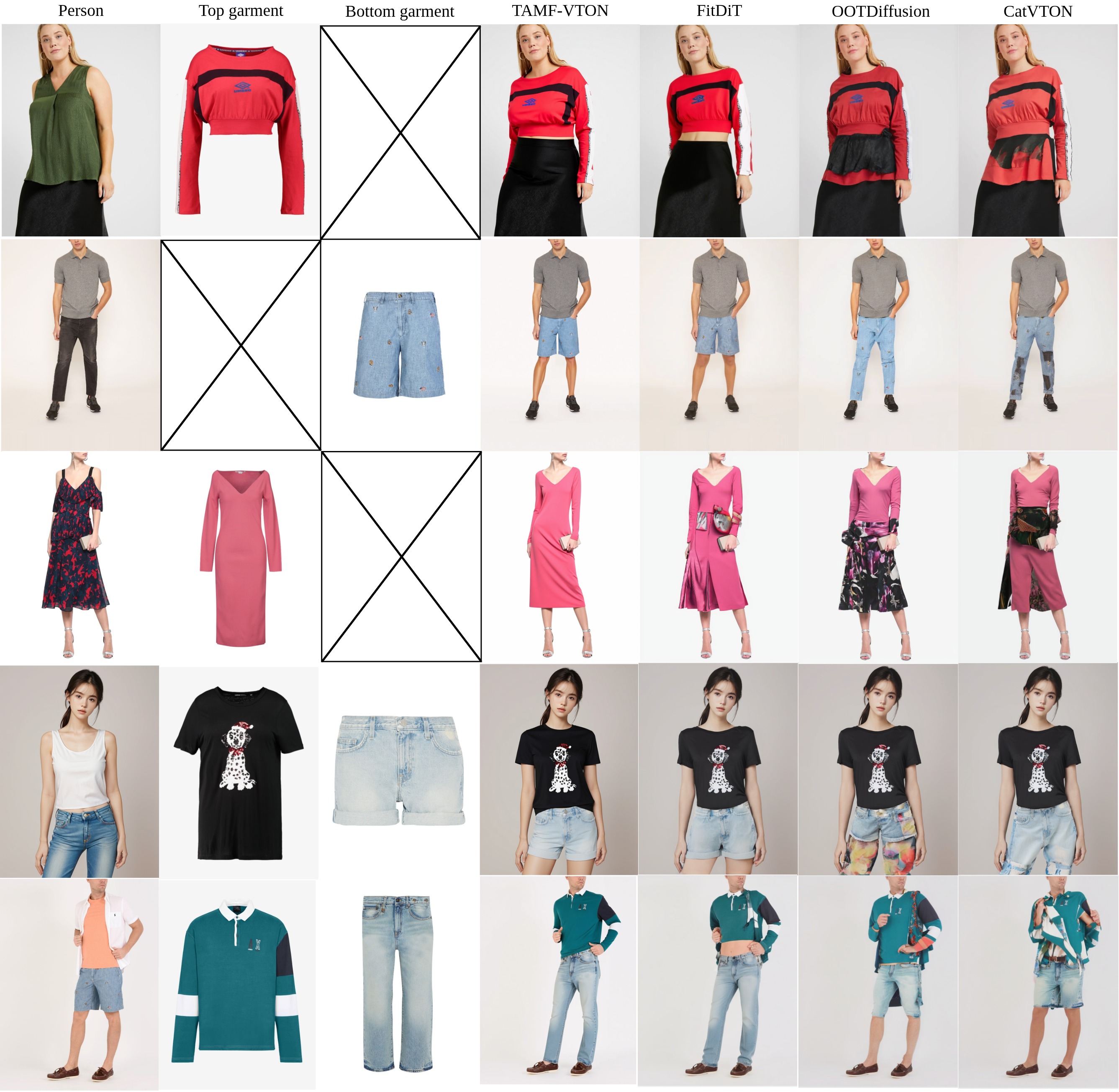}
    \caption{Qualitative comparison on VITON-HD and DressCode. TAMF-VTON preserves garment shape, body structure, and fine-grained texture details more faithfully than recent state-of-the-art methods. Best viewed zoomed in.}
    \label{fig6}
\end{figure*}

\begin{figure*}
    \centering
    \includegraphics[width=0.9\textwidth]{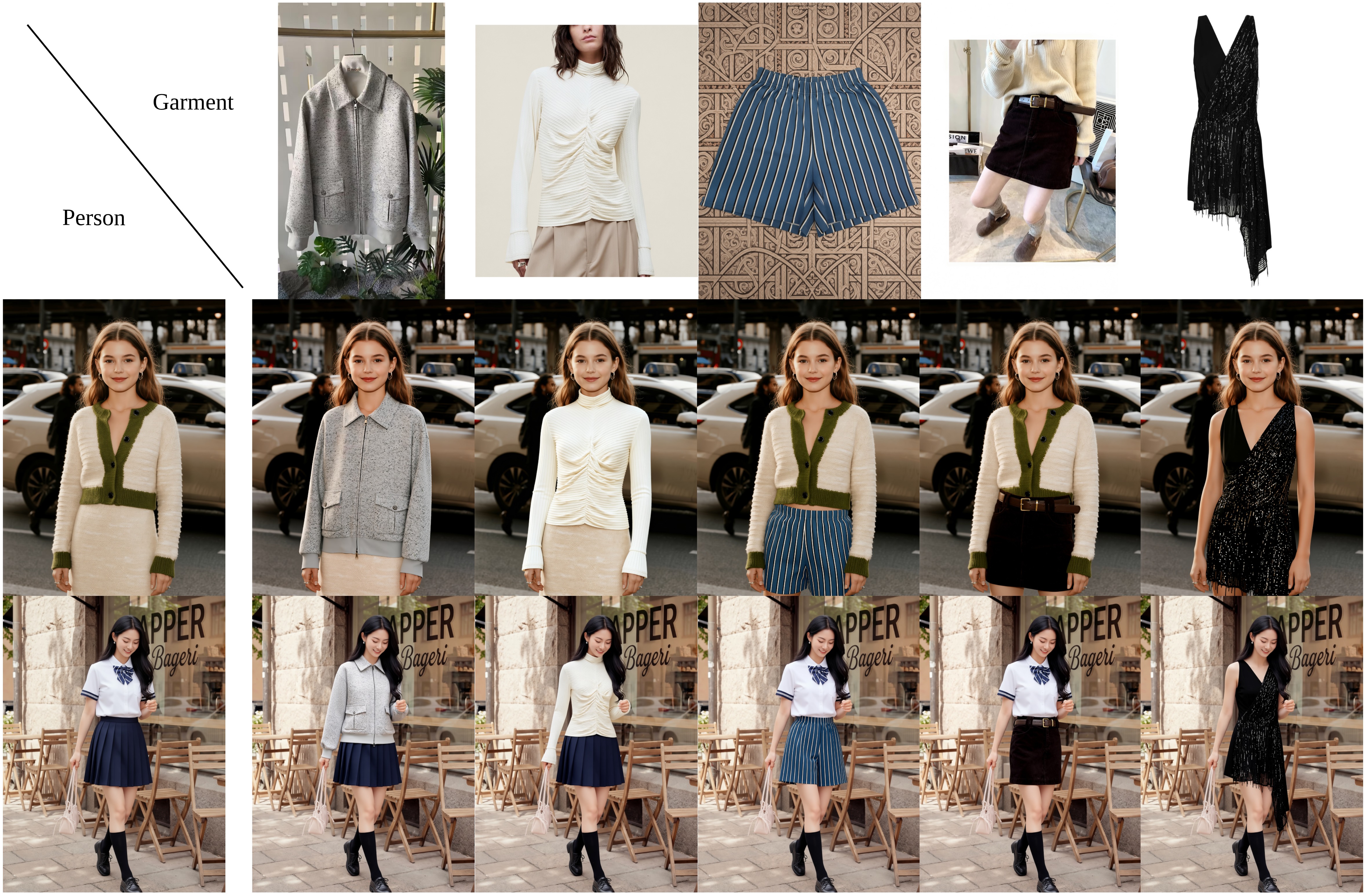}
    \caption{Robust single-garment virtual try-on under complex backgrounds and subject-to-subject transfer settings. Best viewed when zoomed in.}
    \label{fig9}
\end{figure*}

\begin{figure*}
    \centering
    \includegraphics[width=0.9\textwidth]{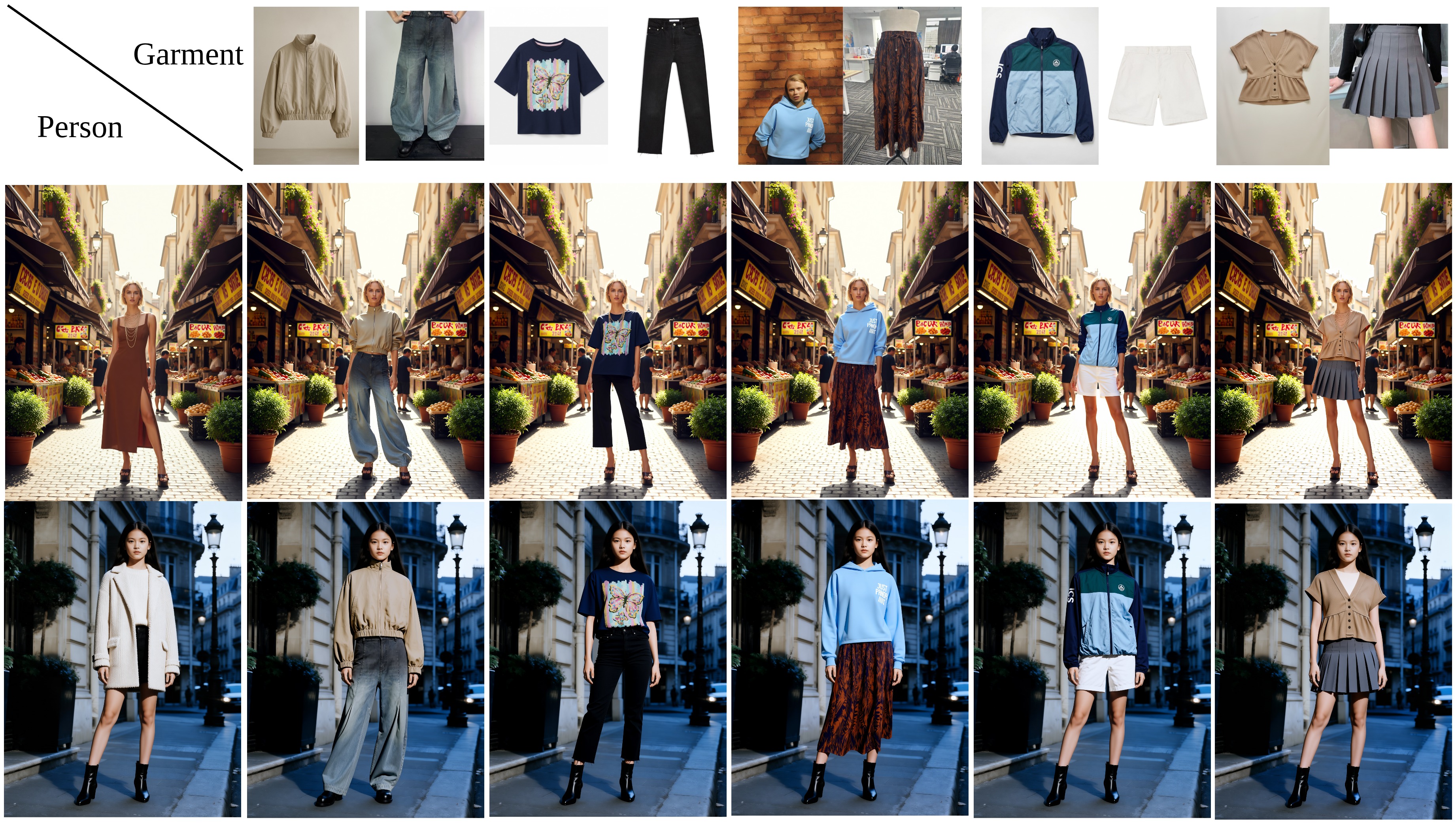}
    \caption{Robust multi-garment virtual try-on under complex backgrounds and subject-to-subject transfer settings. Best viewed when zoomed in.}
    \label{fig10}
\end{figure*}

Figure~\ref{fig8} compares TAMF-VTON with and without the proposed frequency-domain supervision. Without this term, the model tends to generate oversmoothed or distorted local patterns, especially for garments with dense textures or fine fabric structures. In contrast, our full model better preserves high-frequency details, producing textile patterns that more faithfully match the reference garment. Quantitative results in Table~\ref{tab1} further show that frequency-domain supervision brings consistent improvements across evaluation metrics, confirming its effectiveness in enhancing texture fidelity beyond standard flow-matching training.

Additional ablation study on Top-\(k\) expert routing and frequency-domain supervision are presented in the supplementary material.

\section{Limitations and Future Work}
TAMF-VTON achieves high-quality mask-free virtual try-on with flexible multi-garment composition, but several challenges remain in extreme real-world scenarios. Under severe illumination variations, such as overexposed or underexposed conditions, the synthesized garment color may slightly deviate from the reference appearance. In addition, when previously occluded body regions become exposed after try-on, the generated skin details may differ from the target person due to the strong generative prior of the base model. These limitations remain common challenges for current virtual try-on systems. Future work will focus on illumination-aware color consistency and more identity-preserving synthesis for unconstrained real-world applications.

\section{Conclusion}

We present TAMF-VTON, a texture-aware, mask-free framework that formulates virtual try-on as a task-specialized image editing problem. By integrating a visual-context-aware Mixture-of-Experts adaptation scheme, frequency-domain supervision, and an adaptive data curation strategy, our method addresses the coupled challenges of garment deformation, texture fidelity, and identity preservation without relying on error-prone segmentation masks. Extensive experiments demonstrate that TAMF-VTON achieves superior perceptual quality and competitive quantitative performance compared with state-of-the-art methods, while supporting flexible multi-garment composition with efficient inference. 



\bibliographystyle{ACM-Reference-Format}
\bibliography{bibliography}

\clearpage

\end{document}


\title{Supplementary Material for TAMF-VTON}



\maketitle

\section{Necessity of Adaptive Inpainting Strategy}

\begin{figure*}
    \centering
    \includegraphics[width=1.0\textwidth]{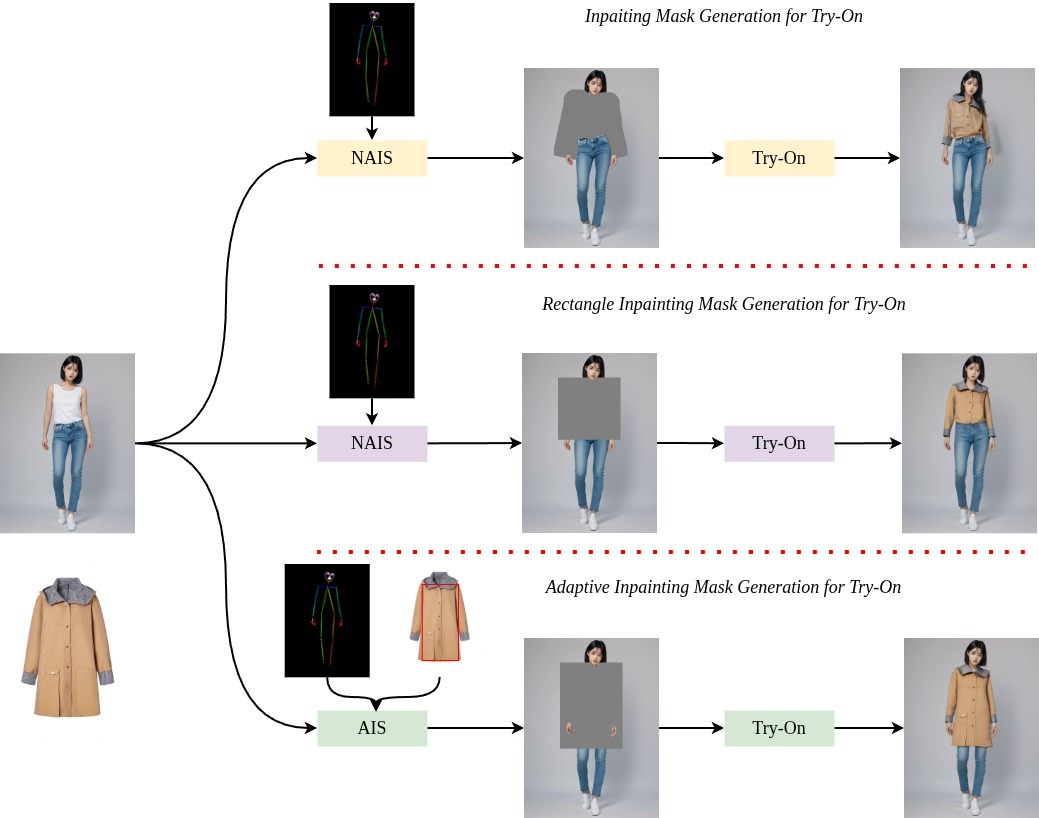}
    \caption{Comparison between non-adaptive inpainting strategies (NAIS) and the proposed adaptive inpainting strategy (AIS). Existing pipelines construct inpainting masks directly from the target person's original clothing regions, resulting in strong coupling between the mask geometry and the source garment. Consequently, they often fail under cross-category or cross-length garment transfer, such as replacing a short upper garment with a long coat. In contrast, AIS adaptively estimates the inpainting region according to both body keypoints and the reference garment geometry, enabling flexible and accurate mask-free try-on across diverse garment styles and lengths.}
    \label{suppl0}
\end{figure*}

Figure~\ref{suppl0} compares the proposed adaptive inpainting strategy (AIS) with two representative non-adaptive inpainting strategies (NAIS). The upper pipeline corresponds to the commonly adopted paradigm used in methods such as OOTDiffusion~\cite{xu2025}, CatVTON~\cite{chong2025}, IDM-VTON~\cite{choi2024}, as well as the predefined masks provided in standard benchmarks. This paradigm is also adopted in the training data synthesis pipelines of prior mask-free approaches~\cite{zhang_boow_vton, wan_11375578, Du_2025_ICCV}. Specifically, the inpainting mask is constructed from the original garment region together with dilated body-part regions derived from human parsing and pose estimation. Since the mask is directly determined by the clothing originally worn by the target person, it remains tightly coupled with the original garment geometry. As a result, the synthesized samples inherit strong geometric priors from the source clothing.

The middle pipeline enlarges the original inpainting region into a rectangular mask, as adopted in FitDiT~\cite{jiang2024}. Although this partially relaxes the spatial constraint, the generated mask still lacks explicit adaptation to the geometry of the reference garment and therefore cannot reliably support large cross-style or cross-length garment transfer.

In contrast, the proposed AIS explicitly models the relationship between the inpainting region and the reference garment. Specifically, we adaptively estimate the target garment extent according to body keypoints and the aspect ratio of the reference garment region, allowing the synthesized inpainting mask to dynamically adjust to different garment lengths and silhouettes. Consequently, AIS naturally supports flexible transfer across diverse garment categories and styles.

As shown in Figure~\ref{suppl0}, when replacing the short upper garment worn by the target person with a long coat, both NAIS pipelines produce geometrically inconsistent try-on results because the inpainting regions remain constrained by the geometry of the original short garment. In contrast, AIS generates a spatially compatible inpainting region and produces a realistic try-on result. These comparisons further validate the effectiveness and flexibility of the proposed adaptive inpainting strategy for mask-free single- and multi-garment virtual try-on.

\section{Experimental Details}
\subsection{Evaluation Metrics}

For quantitative evaluation, we adopt a standard set of metrics, including LPIPS (Learned Perceptual Image Patch Similarity) for perceptual similarity, SSIM (Structural Similarity Index) for structural consistency, and FID (Fréchet Inception Distance) and KID (Kernel Inception Distance) for measuring distribution-level realism with respect to real try-on images. These metrics jointly evaluate perceptual fidelity, structural preservation, and synthesis realism under both paired and unpaired settings.

\subsection{Qualitative Comparison with Strong Baselines}

The main paper already presents comparisons with recent open-source state-of-the-art mask-based virtual try-on methods, including FitDiT~\cite{jiang2024}, OOTDiffusion~\cite{xu2025}, and CatVTON~\cite{chong2025}. 

In the supplementary material, we further compare TAMF-VTON with recent strong multimodal image editing models, including Qwen-Edit-2509~\cite{wu2025}, Flux2-klein-9b~\cite{flux-2-2025}, and Nano Banana Pro~\cite{geminiteam2025geminifamilyhighlycapable}, which natively support multi-image editing and compositional generation.
These general-purpose editing models represent recent advances in multimodal image editing and have demonstrated strong capabilities on virtual try-on tasks.

\subsubsection{Single-garment Try-on in Standard Benchmarks}

\begin{figure*}
    \centering
    \includegraphics[width=0.9\textwidth]{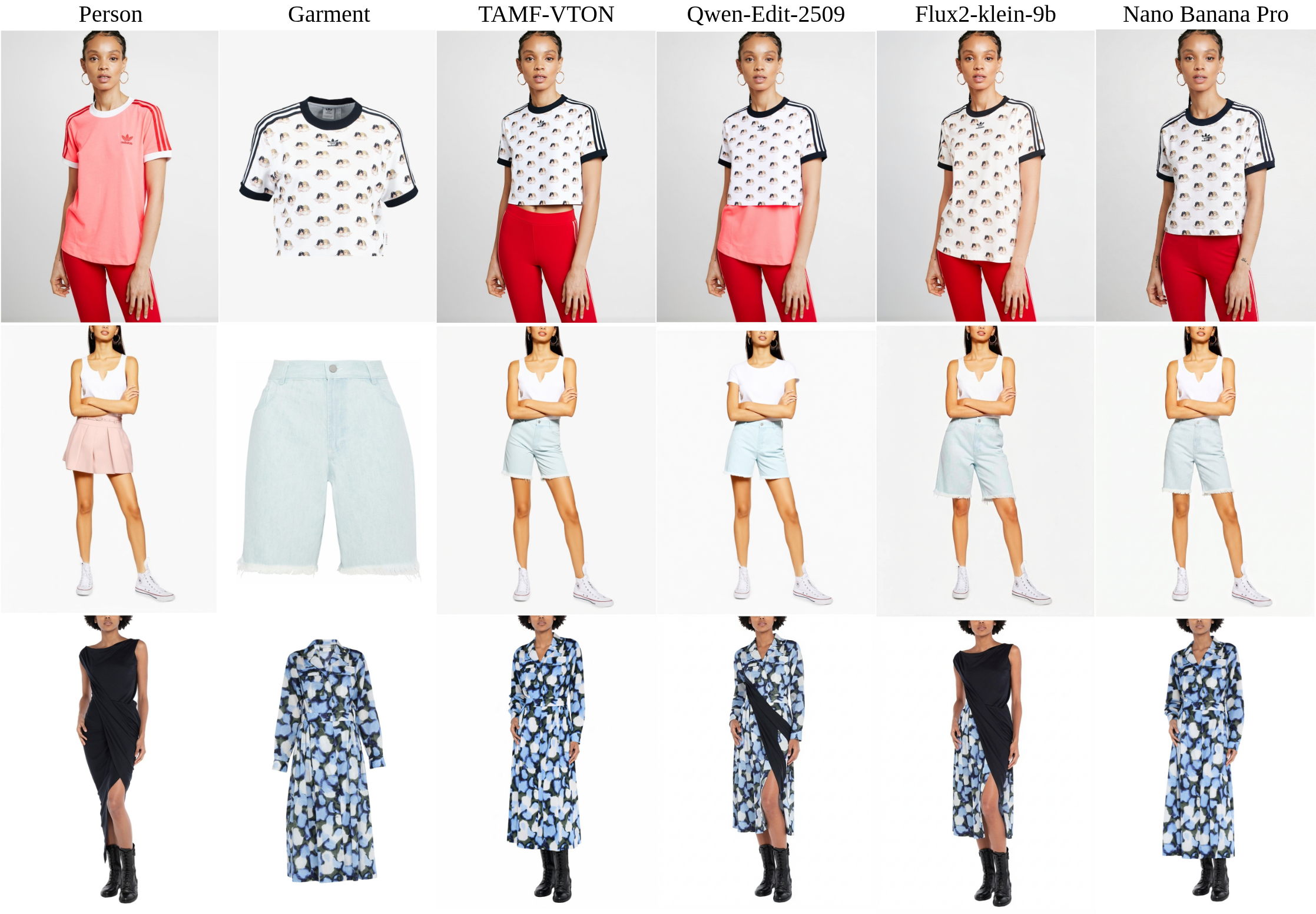}
    \caption{Qualitative comparison on VITON-HD and DressCode. Single-garment try-on results for upper-body, lower-body, and dress categories, transferring flat-lay garments onto studio-captured persons with clean backgrounds. Best viewed zoomed in.}
    \label{suppl1}
\end{figure*}

As shown in Figure~\ref{suppl1}, TAMF-VTON produces more realistic single-garment try-on results from flat-lay garment images to studio-captured persons, outperforming Qwen-Edit-2509 and Flux2-klein-9b, while achieving performance comparable to Nano Banana Pro. In the first row, when transferring a short T-shirt onto a person originally wearing a long T-shirt, Flux2-klein-9b incorrectly extends the reference garment into a long style, while Qwen-Edit-2509 fails to completely remove the original clothing. In the second row, Qwen-Edit-2509 shows weaker preservation of garment details, and Flux2-klein-9b produces overly smooth textures on the shorts. In contrast, TAMF-VTON better preserves fine-grained material and texture details. In the third row, both Qwen-Edit-2509 and Flux2-klein-9b fail to generate coherent dress try-on results, whereas TAMF-VTON produces visually plausible and structurally consistent outputs.

For fair comparison, we use simple natural-language prompts for all editing models, e.g., ``The model in image 1 is dressed in the top garment from image 2'' for upper-body try-on, with similar prompts for lower-body garments and dresses.

\subsubsection{Multi-garment Try-on in Standard Benchmarks}

\begin{figure*}
    \centering
    \includegraphics[width=1.0\textwidth]{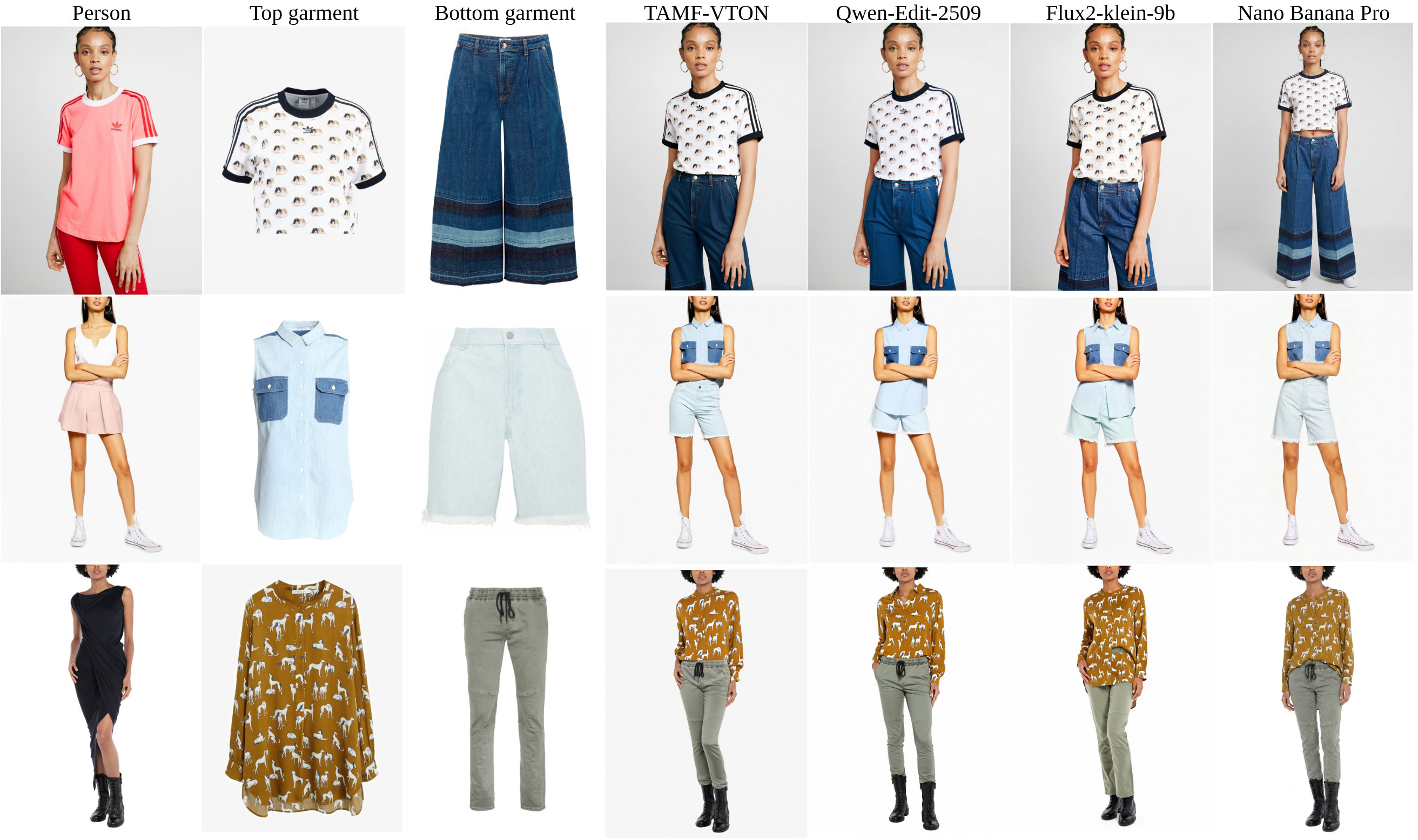}
    \caption{Qualitative comparison on VITON-HD and DressCode. Multi-garment try-on results transferring multiple flat-lay garments onto studio-captured persons with clean backgrounds. Best viewed zoomed in.}
    \label{suppl2}
\end{figure*}

Figure~\ref{suppl2} presents qualitative comparisons for multi-garment try-on from multiple flat-lay garments to studio-captured persons. TAMF-VTON consistently generates coherent and realistic results under multi-garment settings.

In the first row, Qwen-Edit-2509 and Flux2-klein-9b successfully transfer the garments but produce overly smoothed textures, while Nano Banana Pro shows inconsistent spatial alignment in body pose. In the second row, Qwen-Edit-2509 and Flux2-klein-9b generate inconsistent garment boundaries, particularly around the hemline regions. In the third row, all baselines exhibit noticeable pose inconsistency, while Flux2-klein-9b further fails to correctly transfer the upper-body garment.

Overall, TAMF-VTON demonstrates better structure preservation, stronger texture fidelity, and more robust performance in standard multi-garment scenarios compared with existing strong baselines.

\subsubsection{Qualitative Results in Challenging Scenarios}

\begin{figure*}
    \centering
    \includegraphics[width=1.0\textwidth]{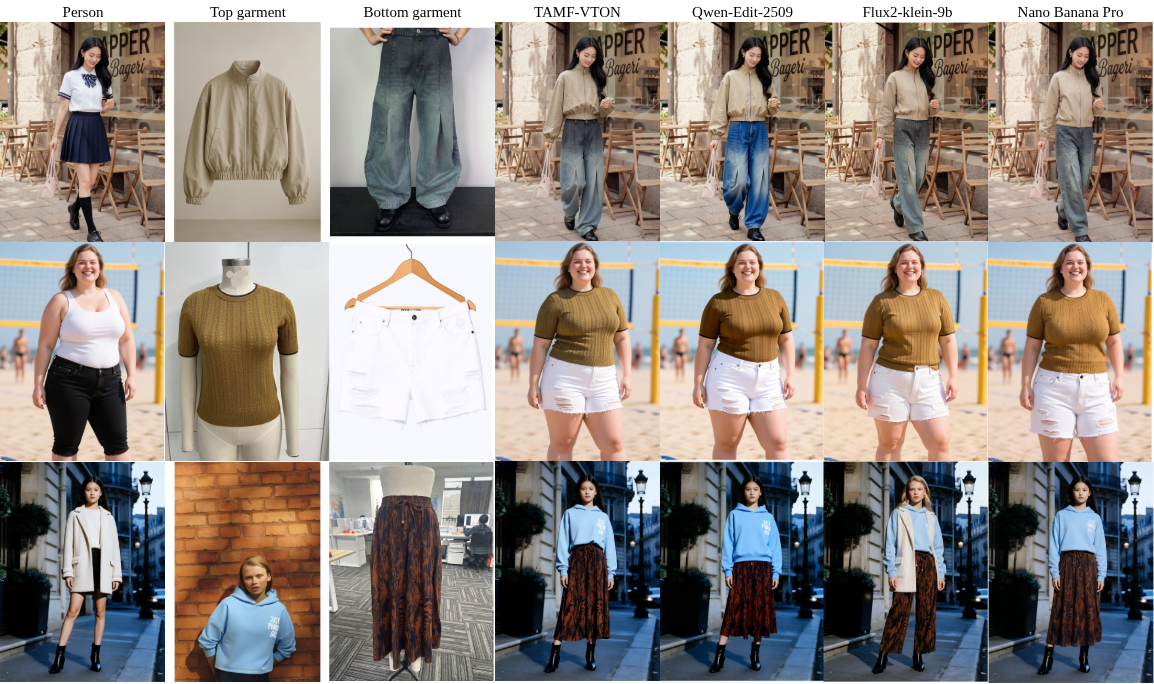}
    \caption{Qualitative comparisons in challenging real-world scenarios, including subject-to-subject transfer, subjects with challenging body-shape variations, and extreme lighting conditions. Compared with recent strong baselines, TAMF-VTON better preserves garment geometry, texture fidelity, subject identity, body structure, and scene consistency under complex conditions. Best viewed when zoomed in.}
    \label{suppl3}
\end{figure*}

We further compare TAMF-VTON with recent strong baselines under several highly challenging real-world scenarios, including subject-to-subject transfer, challenging body-shape variations, and extreme lighting conditions. As shown in Figure~\ref{suppl3}, TAMF-VTON consistently produces more stable and realistic results while better preserving garment geometry, texture fidelity, subject identity, and body structure.

\begin{figure*}
    \centering
    \includegraphics[width=0.9\textwidth]{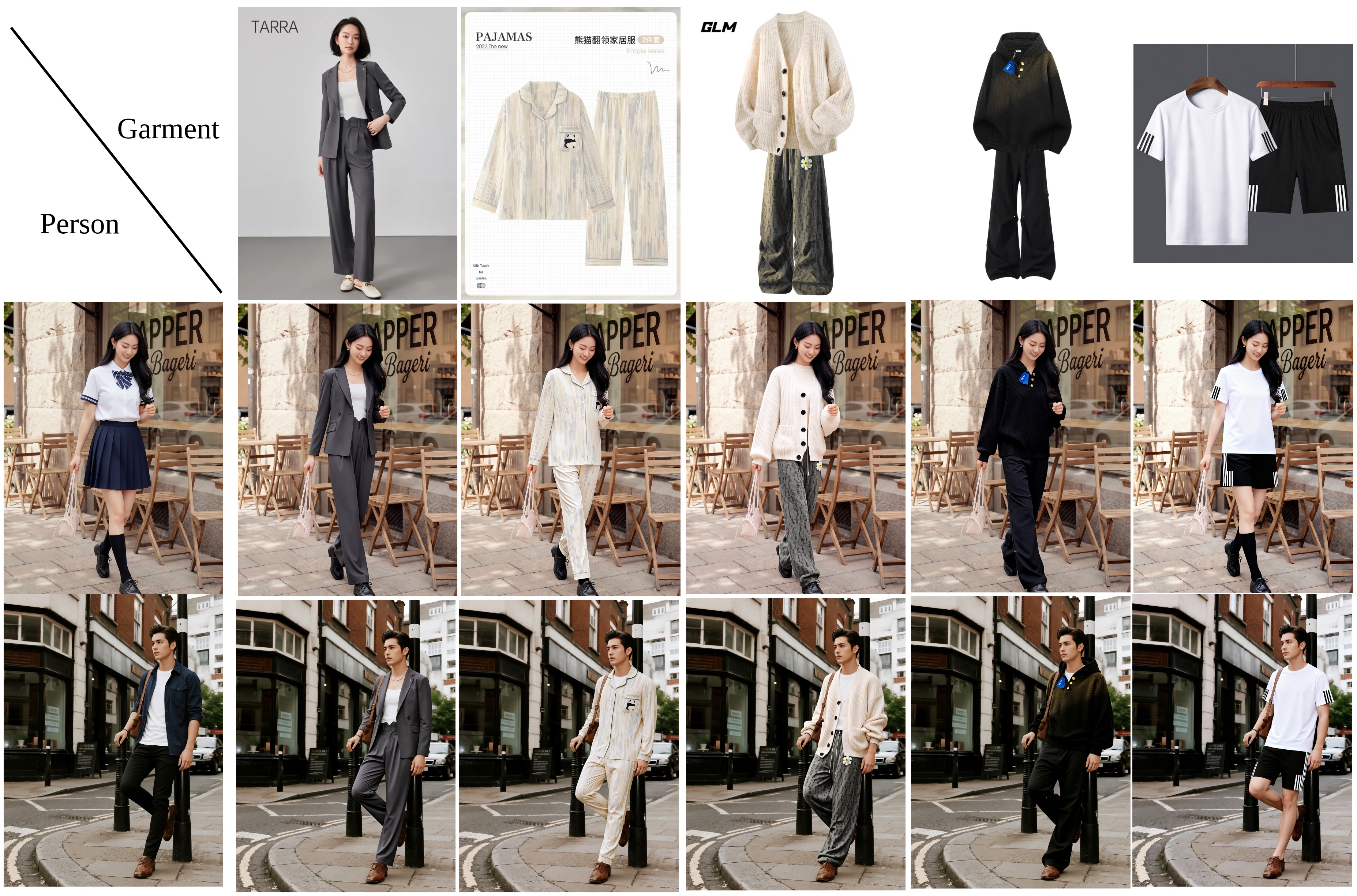}
    \caption{Additional qualitative results of TAMF-VTON under challenging real-world scenarios, with a particular focus on multi-garment and outfit try-on. The results demonstrate that our method maintains strong robustness across diverse and complex settings. Best viewed when zoomed in.}
    \label{suppl4}
\end{figure*}

In the first row, which involves subject-to-subject transfer with loose-fitting pants, Qwen-Edit-2509 fails to preserve the original texture and color distribution, while Flux2-klein-9b and Nano Banana Pro incorrectly reshape the loose pants into a slim-fit style, significantly altering the intended garment geometry. In the second row, which evaluates try-on performance under challenging body-shape variations, Qwen-Edit-2509 generates overly smoothed garment textures, Flux2-klein-9b produces inconsistent upper-garment boundaries, and Nano Banana Pro noticeably changes the pose and body structure of the target person. All three baselines exhibit slight inconsistencies in subject identity preservation. The third row presents examples under extreme lighting conditions. Qwen-Edit-2509 alters both garment color and hemline structure, while Flux2-klein-9b fails to remove the original upper garment and incorrectly transforms the reference skirt into pants. More critically, Flux2-klein-9b exhibits significant subject identity drift. In contrast, TAMF-VTON achieves more reliable and visually coherent results across all challenging scenarios, demonstrating stronger robustness and practical usability in real-world VTON settings. Additional qualitative results demonstrating the robustness of TAMF-VTON in challenging scenarios are shown in Figure~\ref{suppl4}.

\subsection{Quantitative Comparison with Strong Baselines}

\begin{table*}
    \caption{Quantitative results on VITON-HD \cite{choi2021} and DressCode \cite{morelli2022} datasets. We present paired and unpaired evaluation results. Our proposed method consistently outperforms most of the baselines. Bold indicates the best metrics in this table.}
    \label{tab1}
    \centering
    \resizebox{1.0\textwidth}{!}{
    \begin{tabular}{ccccccccccccccccc}
        \toprule
         \multirow{3}{*}{Methods} &  &  \multicolumn{7}{c}{VITON-HD \cite{choi2021}} &  & \multicolumn{7}{c}{DressCode \cite{morelli2022}}\\
         \cline{3-9}
         \cline{11-17}
         &  &  \multicolumn{4}{c}{Paired}&  &  \multicolumn{2}{c}{Unpaired}&  & \multicolumn{4}{c}{Paired}& & \multicolumn{2}{c}{Unpaired}\\
         \cline{3-6}
         \cline{8-9}
         \cline{11-14}
         \cline{16-17}
 & & SSIM $\uparrow$ & LPIPS $\downarrow$ & FID $\downarrow$ & KID $\downarrow$ & & FID $\downarrow$ & KID $\downarrow$ & & SSIM $\uparrow$ & LPIPS $\downarrow$ & FID $\downarrow$ & KID $\downarrow$ & & FID $\downarrow$ &KID $\downarrow$\\
    \hline
         Qwen-Edit-2509 & & 0.793 & 0.217 & 6.640 & 0.502 & & 8.79 & 1.148 & & 0.837 & 0.183 & 3.546 & 1.994 & & 4.916 & 1.732  \\
         Flux2-klein-9b & & 0.780 & 0.236 & 7.325 & 0.853 & & 9.818 & 0.874 & & 0.822 & 0.182 & 6.814 & 7.320 & & 4.494 & 2.190  \\
         Nano Banana Pro & & 0.810 & 0.152 & 5.632 & \textbf{0.129} & & 7.419 & 0.441 & & 0.865 & 0.119 & 10.705 & 1.086 & & 3.672 & \textbf{0.516} \\
         \midrule
         TAMF-VTON &  &  \textbf{0.913} &  \textbf{0.052} &  \textbf{4.322} &  0.277&  &  \textbf{6.268} &  \textbf{0.265} &  & \textbf{0.933} & \textbf{0.040} & \textbf{2.363}& \textbf{0.360}& & \textbf{3.321} & 0.597\\ 
         \bottomrule
    \end{tabular}
    }
    
\end{table*}

Table~\ref{tab1} reports quantitative results on VITON-HD and DressCode under both paired and unpaired settings. Overall, TAMF-VTON achieves the best performance on the majority of metrics across both datasets, demonstrating consistent improvements in structural fidelity and perceptual quality.

In the paired setting, our method consistently improves SSIM while substantially reducing LPIPS, indicating better preservation of garment structure and fine-grained texture details. In addition, TAMF-VTON achieves lower FID and competitive KID scores, reflecting improved global realism and distribution alignment with ground-truth images.

In the unpaired setting, where evaluation relies on distribution-level similarity rather than pixel-aligned supervision, TAMF-VTON consistently achieves the lowest FID and KID scores on VITON-HD, while remaining highly competitive on DressCode. These results demonstrate strong generalization ability, particularly in cross-category try-on scenarios.

Compared with strong baselines such as Qwen-Edit-2509, Flux2-klein-9b, and Nano Banana Pro, our method provides a more balanced performance across different metrics, avoiding the common trade-off between structural consistency and texture fidelity. These results further demonstrate the effectiveness of our MoE-based adaptation and frequency-domain supervision for high-fidelity mask-free virtual try-on.

\section{Ablation Study}

\subsection{Effect of Top-\(k\) Expert Routing}

We further investigate the influence of the top-\(k\) routing strategy in the proposed MoE adaptation module. Specifically, we vary the number of activated experts for each multimodal token while keeping all other training settings unchanged. Table~\ref{topk} reports quantitative results under the unpaired evaluation protocol. We adopt the unpaired setting because it better reflects practical mask-free VTON scenarios, where garments are transferred across different styles, lengths, and categories. In this setting, FID and KID better capture distribution-level realism and generalization capability.

Overall, \(k=2\) provides the most balanced performance across both datasets. Compared with \(k=1\), activating two experts significantly improves both FID and KID, indicating that a single adaptation path is insufficient for jointly modeling the coupled objectives of garment deformation, texture preservation, and identity consistency. Although \(k=3\) achieves the best KID score on VITON-HD, its FID slightly deteriorates and its performance on DressCode becomes less stable, suggesting weaker distribution alignment across datasets.

When \(k=4\), all experts are activated simultaneously, making the sparse MoE adaptation behave similarly to a dense LoRA-style formulation with limited expert specialization. As a result, both FID and KID degrade noticeably on the two datasets. Considering both quantitative performance and computational efficiency, we adopt top-\(2\) routing in all experiments.

\begin{table}
    \caption{Ablation study on the top-\(k\) routing strategy in the proposed MoE adaptation module. We vary the number of activated experts for each multimodal token and report quantitative results on unpaired VITON-HD and DressCode. The results show that \(k=2\) achieves the best trade-off between synthesis quality and computational efficiency.}
    \label{topk}
    \centering
    \begin{tabular}{ccccccc}
        \toprule
          \multirow{2}{*}{Top-\(k\)} & & \multicolumn{2}{c}{VITON-HD} & & \multicolumn{2}{c}{DressCode} \\
          \cline{3-4}
          \cline{6-7}
          & & FID $\downarrow$ & KID $\downarrow$ & & FID $\downarrow$ & KID $\downarrow$ \\
          \hline
          \(k=1\) & & 7.201 & 0.332 & & 4.260 & 0.763 \\
          \(k=2\) & & \textbf{6.268} & 0.265 & & \textbf{3.321} & \textbf{0.597} \\
          \(k=3\) & & 6.432 & \textbf{0.254} & & 3.541 & 0.663 \\
          \(k=4\) & & 8.010 & 0.458 & & 4.886 & 0.893 \\
          
        \bottomrule
    \end{tabular}
\end{table}

\subsection{Ablation Study on Frequency-domain Supervision}

\begin{figure*}
    \centering
    \includegraphics[width=1.0\textwidth]{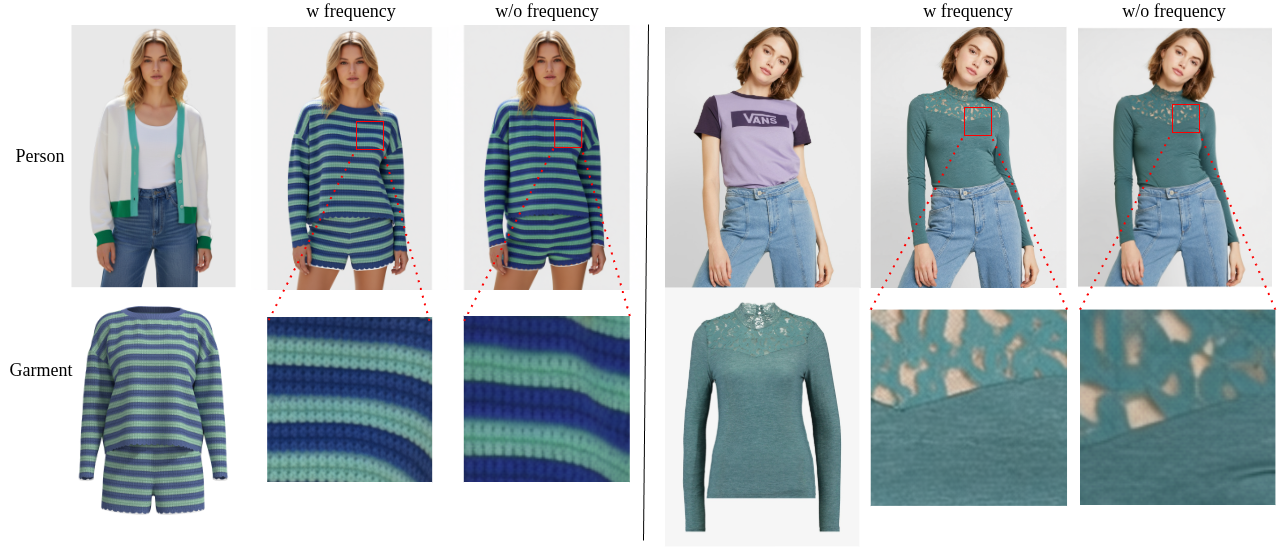}
    \caption{Ablation study on the proposed frequency-domain supervision. \textit{w frequency} denotes the full model with frequency-domain supervision, while \textit{w/o frequency} removes the spectral supervision term. The proposed supervision significantly improves the preservation of fine-grained fabric textures and material appearance. Best viewed when zoomed in.}
    \label{suppl5}
\end{figure*}

We conduct additional experiments to further validate the effectiveness of the proposed frequency-domain supervision for preserving fine-grained garment textures and material appearance. As shown in Figure~\ref{suppl5}, the variant with frequency-domain supervision is denoted as \textit{w frequency}, while the variant without it is denoted as \textit{w/o frequency}. All other training and inference settings remain identical.

In the left example, a knitted outfit is transferred onto the target person, while the right example involves a cotton-linen upper garment. As highlighted by the red boxes, frequency-domain supervision improves the preservation of fine fabric textures and material appearance. In contrast, the variant without frequency-domain supervision tends to generate overly smooth and blurred garment textures.

\clearpage

\bibliographystyle{ACM-Reference-Format}
\bibliography{bibliography}